\documentclass[]{fairmeta}

\usepackage{amsmath}
\usepackage{amssymb}
\usepackage{mathtools}
\usepackage{amsthm}
\usepackage{comment}
\usepackage{siunitx}
\usepackage{pifont}
\usepackage{adjustbox}
\usepackage{makecell}
\usepackage{wrapfig}
\usepackage{algorithm}
\usepackage{algorithmicx}
\usepackage{algpseudocode}
\usepackage{listings}
\usepackage{manyfoot}
\usepackage[title]{appendix}
\usepackage{mathrsfs}
\usepackage{colortbl}

\newcommand{\cmark}{\ding{51}} 
\newcommand{\xmark}{\ding{55}} 

\newcommand{\na}{\textemdash}
\newcommand{\drop}[1]{\textcolor{gray}{$\downarrow$#1}}

\theoremstyle{plain}

\theoremstyle{definition}

\theoremstyle{remark}

\title{InstructSAM: Segment Any Instance with Any Instructions}

\author[1,*]{Yuqian Yuan}
\author[2,*,\dagger]{Wentong Li}
\author[1,*]{Zhaocheng Li}
\author[1,*]{Yutong Lin}
\author[1]{Juncheng Li}
\author[1]{Siliang Tang}
\author[1]{Jun Xiao}
\author[1]{Yueting Zhuang}
\author[1,\ddagger]{Wenqiao Zhang}

\affiliation[1]{Zhejiang University}
\affiliation[2]{Nanjing University of Aeronautics and Astronautics}

\contribution[*]{Equal contribution}
\contribution[\dagger]{Project lead}
\contribution[\ddagger]{Corresponding author}

\abstract{In this paper, we introduce \textit{InstructSAM}, a unified and   streamlined framework designed for robust multi-instance segmentation under arbitrary instructions. We formulate instruction-driven instance segmentation as a set-structured query prediction problem and propose an explicit reasoning-to-instance query interface that  elegantly bridges a vision-language model (VLM) and SAM3. Specifically, a bank of learnable instance queries is injected into the VLM and contextualized with instruction and visual information, enabling each query to serve as an instance-aware slot. A hybrid-attention mechanism further promotes interaction among these queries, visual tokens, and instruction tokens, improving instance enumeration and reducing duplicate predictions.
The resulting LLM-conditioned queries are projected into SAM3’s detector query space to drive  accurate multi-instance segmentation in a single forward pass. This design equips SAM3 with high-level instruction understanding, compositional reasoning, and instance-level set prediction without modifying its core architecture. To support training and evaluation, we  further construct \texttt{Inst$^{2}$Seg}, a high-quality and large-scale instruction-based instance segmentation dataset and benchmark that couples free-form instructions with instance-level masks. Extensive experiments show that only 2B-scale InstructSAM achieves strong results across complex instruction-driven and phrase-level referring segmentation benchmarks, outperforming prior end-to-end methods and SAM3’s agentic pipeline while enabling efficient single-pass multi-instance prediction.

\hspace*{0.5em}
\begin{tabular}{@{}l@{\hspace{1.0em}}l@{}}
    \makebox[1.5em][l]{\raisebox{-0.2em}{\includegraphics[height=1.1em]{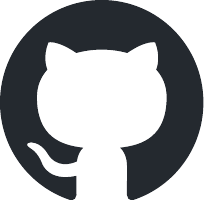}}}
    &
    \textbf{GitHub:} \url{https://github.com/DCDmllm/InstructSAM}
    \\
    \makebox[1.5em][l]{\raisebox{-0.2em}{\includegraphics[height=1.1em]{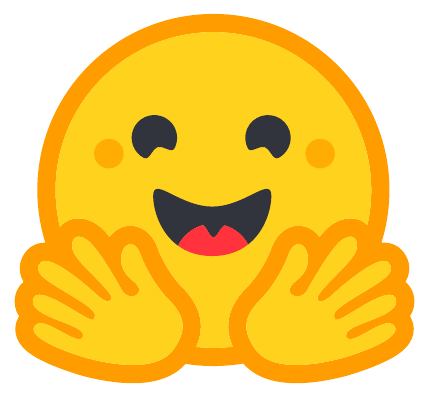}}}
    &
    \textbf{Dataset:} \url{https://huggingface.co/datasets/CircleRadon/Inst2Seg}
    \\

    \makebox[1.5em][l]{\raisebox{-0.2em}{\includegraphics[height=1.1em]{assets/hf-logo.pdf}}}
    &
    \textbf{Model:} \url{https://huggingface.co/CircleRadon/InstructSAM-2B/}
\end{tabular}

}

\begin{document}

\maketitle
\section{Introduction}\label{sec1}

\begin{figure*}[t]
\centering
\includegraphics[width=\linewidth]{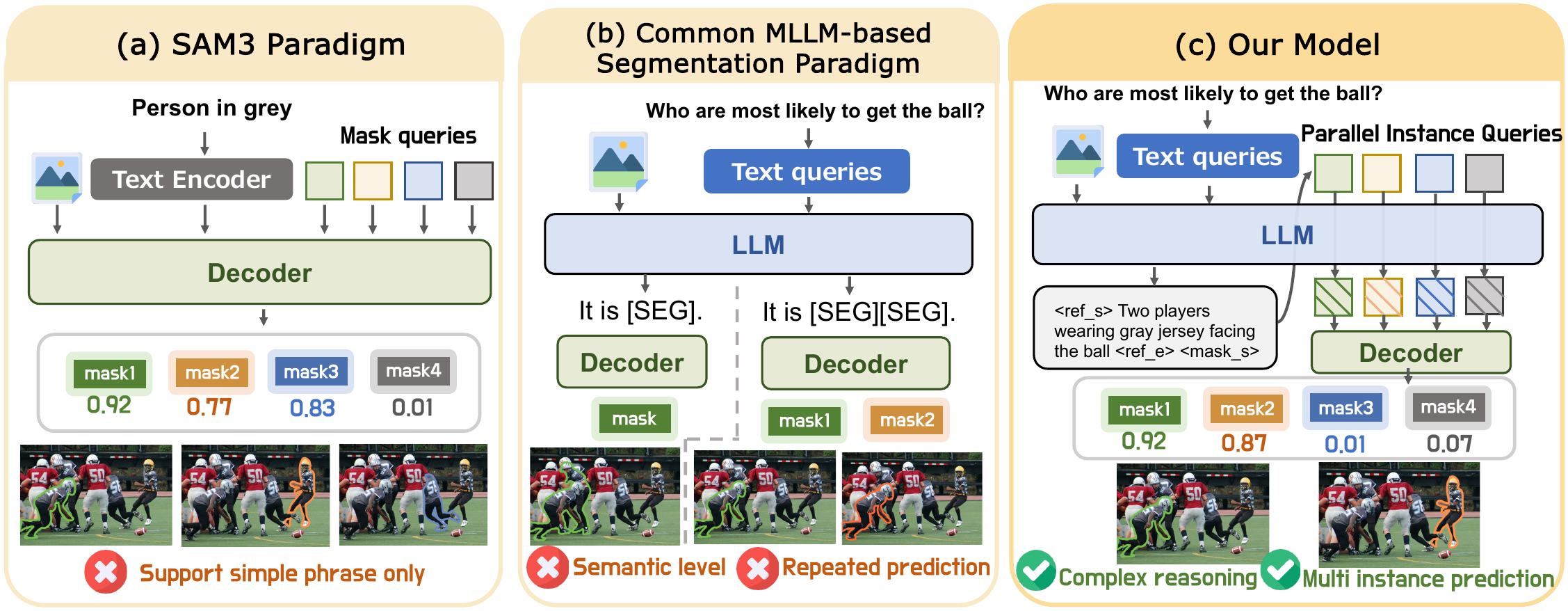}
\vspace{-5mm}
\caption{Comparison of instruction-driven segmentation paradigms: (a) SAM3 handles concept-level prompts but struggles with complex instructions; (b) MLLM-based mask-token generation yields semantic masks or inconsistent multi-instance results; (c) our method uses LLM reasoning to condition an explicit set of instance-aware object queries that guide SAM3 for coherent multi-instance segmentation.}\label{fig: intro}
\vspace{-3mm}
\end{figure*}

Segmenting objects in images and videos is a foundational capability for embodied agents~\cite{dang2025rynnec,yuan2025eoc,xin2026agentvln}, autonomous perception~\cite{li2024transformer}, healthcare~\cite{lin2025healthgpt,xie2025heartcare,lin2026omnict} and visual edition~\cite{yuan2026lmms,zhong2026unified}. The line work of Segment Anything Model (SAM) has significantly advanced this direction by enabling promptable segmentation with strong generalization~\cite{kirillov2023segment,ravi2024sam,carion2025sam}. In particular, the recent SAM3~\cite{carion2025sam} extends promptable segmentation to open-world concept-level, multi-instance settings, where a short noun phrase (\textit{e.g.},  ``traffic cone'') can retrieve and segment multiple instances in a scene, as shown in Fig.~\ref{fig: intro}(a). Despite this promising progress, a critical gap remains between concept-level prompting and real-world user intent. In practice, users rarely communicate their targets as isolated noun phrases; instead, they often issue complex, compositional instructions involving attributes (``the small mugs''), spatial constraints (``on the left''), relations (``next to the laptop''), exclusion (``except the one in front''), or counting (``the two largest''). Such instructions require nontrivial semantic parsing, visual reasoning, and instance-level grounding, as the target object set is often implicitly defined instead of explicitly specified by a single concept label.

Existing attempts to handle complex instructions mainly follow two paradigms. One common solution is an agentic decomposition-and-filtering pipeline,
where a large vision-language model (VLM), such as Qwen-VL~\cite{bai2025qwen2} or Gemini~\cite{comanici2025gemini}, rewrites the complex instruction into one or more concept-level prompts,  repeatedly invokes SAM3 to generate candidate masks, and then post-filters the results with heuristics or verification prompts. 
However, this indirect process is slow, brittle, and prone to semantic loss, as rewriting may discard fine-grained constraints and iterative filtering can accumulate errors.
Another line of work equips LLMs with a special segmentation token, \textit{i.e.} \texttt{[SEG]}, whose hidden state is decoded into a mask, as in LISA~\cite{lai2024lisa} and Sa2VA~\cite{yuan2025sa2va}, as illustrated in Fig.~\ref{fig: intro}(b). While effective for reasoning-driven semantic segmentation, this token-as-mask interface is not inherently instance-discriminative.
LISA++~\cite{yang2023lisa++} extends this paradigm by emitting multiple \texttt{[SEG]} tokens for instance prediction. However, because \texttt{[SEG]} is a shared symbol without an explicit instance-binding mechanism, the resulting masks often collapse to duplicates or become unstable, producing repeated or inconsistent outputs. Moreover, autoregressive generation of multiple \texttt{[SEG]} tokens increases inference latency as the number of target instances grows.

In this paper, we propose \textit{InstructSAM}, a unified framework for segmenting arbitrary instances under arbitrary instructions via an explicit reasoning-to-instance interface.
Rather than forcing the LLM to directly ``speak masks'' token by token, 
InstructSAM leverages its general-purpose reasoning capability to interpret complex instructions and translate them into a set-structured, instance-aware query representations.
These representations serve as an explicit interface to SAM3, enabling coherent and efficient multi-instance segmentation.
Concretely, we introduce a bank of learnable queries into the LLM  as parallel instance slots. Through bidirectional interactions among the queries, together with instruction and visual context, these slots are contextualized into instance-specific embeddings that capture potential target instances implied by the instruction.
The resulting LLM-conditioned queries are then projected into SAM3's  detector query space, where they directly drive the detector and mask decoder to localize and segment multiple instances in a single forward pass.
This design bridges instruction reasoning and mask prediction, enabling compositional understanding and coherent instance enumeration, as shown in Fig.~\ref{fig: intro}(c).

To further advance instruction-based instance segmentation, we introduce \textbf{\texttt{Inst$^{2}$Seg}}, 
a large-scale dataset and benchmark that couples free-form instructions with instance-level masks. 
Built through a carefully designed annotation pipeline, \texttt{Inst$^{2}$Seg} contains 500K QA pairs for training and a dedicated benchmark with 3,328 manually verified instructions. The benchmark spans diverse real-world scenarios and instruction types, covering single-target, multi-target, and no-target cases to enables systematic evaluation of coherent instance-level mask prediction under complex instructions.
Extensive experiments demonstrate that the 2B-scale InstructSAM achieves accurate instance-level segmentation under both complex instructions and referring phrases. It significantly outperforms prior state-of-the-art end-to-end approaches and SAM3’s agentic pipeline at the same model scale, while delivering robust performance across scenes with varying object densities and levels of semantic ambiguity.

We summarize our contributions as follows:

\begin{itemize}
\item We present \textit{InstructSAM}, a unified end-to-end framework for instruction-conditioned multi-instance segmentation via an explicit reasoning-to-instance query interface.

\item We introduce a bank of  learnable  queries within the LLM as parallel instance slots, coupled with a hybrid-attention mechanism for coherent, instruction-conditioned set prediction.

\item We construct  \texttt{Inst$^{2}$Seg}, a large-scale instruction-based instance segmentation dataset and benchmark covering single-target, multi-target, and no-target scenarios.

\item Extensive experiments demonstrate that 2B-scale InstructSAM substantially outperforms
prior end-to-end methods and SAM3’s agentic pipeline across established and newly introduced benchmarks.

\end{itemize}

\section{Related Work}\label{sec2}

\subsection{Segment Anything Models}\label{subsec2}

The ``Segment Anything'' line of work has fundamentally reshaped generic visual segmentation by introducing promptable models that generalize across categories and domains. SAM~\cite{kirillov2023segment} formulates segmentation as a prompt-to-mask task, where points, boxes, or coarse masks guide a mask decoder conditioned on image embeddings. Follow-up works~\cite{zhang2023faster,ke2023segment,zhao2023fast} extend SAM along several practical axes, including efficiency and robustness. SAM2~\cite{ravi2024sam} advances the paradigm to videos by introducing memory-based temporal propagation and interactive refinement. 
More recently, SAM3~\cite{carion2025sam} 
broadens 
promptable segmentation to \emph{open-world, concept-level multi-instance} settings, enabling a short noun phrase to retrieve and segment multiple object instances.
This capability significantly improves usability in multi-object scenes, yet it still primarily relies on concise concept prompts and is not designed to directly handle complex compositional  instructions that require  reasoning, exclusion, or counting. To address this issue, SAM3-I~\cite{li2025sam3} equips SAM3 with instruction-aware adapter and trains it to map natural-language instructions to masks. While promising, this direction typically requires modifying and retraining the segmentation model to internalize instruction understanding. In constrast, our goal is to preserve SAM3 as a strong open-world segmenter and interfacing it with a reasoning-capable VLM through an explicit query-based mechanism.

\subsection{Multi-modal Grounded Segmentation}\label{subsec3}

A growing body of work studies how to endow multi-modal large language models (MLLMs)~\cite{Qwen3-VL,team2023gemini,liu2023visual,yuan2024osprey,yuan2025videorefer,yuan2025pixelrefer,li2025tokenpacker,zhang2024hyperllava,wang2026mau,wang2026crossview,zheng2026iad} with pixel-level grounding, enabling them to respond to free-form instructions with segmentation masks. A dominant design paradigm is the \emph{embedding-as-mask} interface: the MLLM is augmented with a special segmentation token (\textit{e.g.}, \texttt{<SEG>}), whose embedding is projected into the prompt space of a mask decoder (often \textit{SAM-style}) and decoded into a mask in an end-to-end fashion~\cite{lai2024lisa,rasheed2024glamm,ren2024pixellm,yuan2025sa2va,bai2024one,yan2024visa}. These methods align phrase-level semantics with pixel outputs, but most still rely on emitting a segmentation token per semantic grounded region. To move from single-region semantic grounding to \emph{multi-instance} prediction, LISA++~\cite{yang2023lisa++} yields multiple \texttt{<SEG>} tokens for instance segmentation and employs bipartite matching to assign each predicted mask to a ground-truth instance during training. In parallel, X-SAM~\cite{wang2025x} targets a broader ``any segmentation'' formulation by standardizing textual prompts with phrase delimiters. In contrast to directly generating mask tokens in an auto-regressive manner, our InstructSAM leverages the MLLM primarily for instruction-level reasoning and instance enumeration, and interfaces it with SAM3 through an explicit set of instance-aware object queries, enabling coherent and efficient multi-instance segmentation under complex instructions.

\section{Method}\label{sec3}

In this section, we first formulate the task of instruction-driven instance segmentation task. We then present \textbf{InstructSAM}, an instance-aware segmentation framework that follows open-form instructions and predict a set of instance masks. Finally, we detail the training objectives used to optimize the proposed framework.

\subsection{Problem Formulation}\label{subsec4}

Instruction-driven instance segmentation aims to predict \emph{instance-level}  masks from an input image and a free-form natural-language instruction. Formally, given an image $x_{\mathrm{img}}$ and an instruction text $x_{\mathrm{txt}}$, the model  outputs a variable-size \emph{set} of instance masks $\mathcal{M}=\{M_i\}_{i=1}^{N}$, where each $M_i \in \{0,1\}^{H \times W}$ denotes the binary mask of the $i$-th instance satisfying  the instruction, and $N$ is the number of selected instances, which can be zero. 
The instruction $x_{\mathrm{txt}}$ is \emph{open-form}, ranging from a category name (\textit{e.g.}, ``chair'') or a referring phrase (\textit{e.g.}, ``the leftmost chair'') to a complex instruction involving attributes, relations, counting, exclusion, or implicit intent (\textit{e.g.}, ``the objects on the table that should be thrown away''). 
Therefore, a model must jointly perform language understanding, visual grounding, and instance separation under open-vocabulary settings. We formulate this task as a set prediction problem:
\begin{equation}
\begin{aligned}
\mathcal{Y} =  f_{\theta}(x_{\mathrm{img}}, x_{\mathrm{txt}})
 = \big\{(M_i, s_i)\big\}_{i=1}^{N}, \\
\text{s.t. } M_i \in \{0,1\}^{H \times W}, \ s_i \in [0,1], \ N \ge 0 .
\end{aligned}
\end{equation}
Here, $s_i$ is the confidence score of the $i$-th predicted mask, $N$ denotes the number of instances.

\begin{figure*}[t]
\centering
\includegraphics[width=\linewidth]{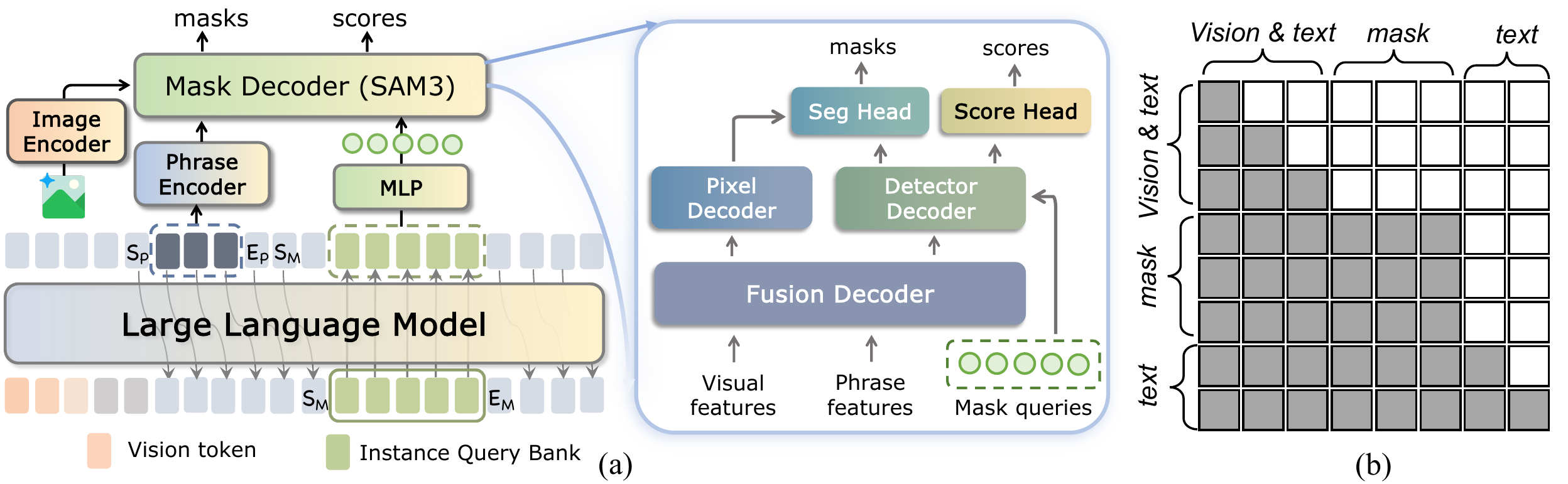}
\caption{\textbf{Overview of the InstructSAM framework.} (a) InstructSAM integrates a multimodal LLM, a set of parallel learnable mask queries, and a mask decoder to generate segmentation masks. (b) Illustration of the hybrid-attention design within the multimodal LLM.}\label{fig: model}
\end{figure*}

Compared with conventional semantic-level reasoning segmentation~\cite{lai2024lisa},
this task is more challenging because it requires not only locating the relevant semantic regions but also separating and enumerating distinct object instances.
It also differs from typical referring segmentation~\cite{li2025sam3}, where the query is usually a concise noun phrase that explicitly specifies the target. In contrast, instruction-driven instance segmentation must handle open-form and  often implicit instructions. 
This capability is essential for embodied perception and robot manipulation, where agents must identify which specific instance to interact with (\textit{e.g.}, ``pick up the mug closest to the sink''), enabling reliable grasp planning, collision avoidance, and sequential decision making.

\subsection{Overview of InstructSAM}\label{subsec5}

As illustrated in Fig.~\ref{fig: model}(a), InstructSAM consists of three components: (\textit{i}) a multimodal LLM $\mathcal{F}$ for instruction understanding, multimodal fusion and instance-slot contextualization; (\textit{ii}) a bank of parallel learnable mask queries $\mathcal{Q}$ that explicitly parameterize \emph{instance slots} as the interface between instruction reasoning and mask prediction; and (\textit{iii}) a set-prediction mask decoder $\mathcal{D}$, instantiated with SAM3, for multi-instance localization and mask decoding.

Crucially, 
we inject a bank of learnable \emph{mask queries} into the LLM  as parallel instance slots, as shown in Fig.~\ref{fig: model}(a).  These queries define an explicit \emph{slot space} where different slots can specialize to different target instances within the same image. Given the instruction, visual features, and textual context produced by the LLM, each learnable query is contextualized into a semantically grounded instance embedding for downstream mask prediction.
To encourage set-level coherence and suppress duplicate predictions, we further design a \emph{hybrid-attention} pattern that allows each instance slot to globally integrate visual evidence, instruction cues, and information from other slots, as illustrated in Fig.~\ref{fig: model}(b). The resulting LLM-conditioned query embeddings are then projected into SAM3’s detector query space and consumed by its detector and mask decoder to produce multiple instance masks in a single forward pass. This architecture enables InstructSAM to combine the reasoning capability of MLLMs with the strong open-world multi-instance segmentation ability.

\subsubsection{Parallel Instance Query Bank}
Given an image $x_{\mathrm{img}}$ and an instruction $x_{\mathrm{txt}}$, the  image encoder produces visual tokens $\mathbf{V} = \mathrm{Enc}_{\mathrm{img}}(x_{\mathrm{img}})\in \mathbb{R}^{L_v\times d}$, while the instruction is tokenized into text embeddings $\mathbf{T}=\mathrm{Emb}(x_{\mathrm{txt}})\in\mathbb{R}^{L_t\times d}$. We introduce a learnable query bank $\mathcal{Q}=\{\mathbf{q}_k\}_{k=1}^{K}$ as parallel instance slots, where $\mathbf{q}_k\in\mathbb{R}^{d}$ and $K$  controls the maximum number of instances that can be predicted in a single forward pass.

A key design of InstructSAM is to replace conventional autoregressively generated segmentation tokens with parallel learnable queries.
Specifically, when the model encounters the trigger token \texttt{<mask\_start>}, we insert the learnable query bank $\mathcal{Q}$ into the multimodal sequence and process it  with the LLM in a \emph{single} forward pass:
\begin{equation}
\mathbf{X} = [\mathbf{V};\mathbf{T};\mathbf{T}_{\text{phrase}};\texttt{<mask\_start>};\mathbf{q}_1;\ldots;\mathbf{q}_K],
\end{equation}
where $\mathbf{T}_{\text{phrase}}$ denotes a short target phrase  (\textit{e.g.}, a concise referring description or resolved target phrase) generated by the LLM to provide auxiliary conditioning for segmentation.
This phrase serves as  a compact, grounded summary of the open-form instruction, which helps stabilize the interface with the mask decoder and reduce ambiguity, especially when the instruction involves implicit intent or multi-step reasoning.

The LLM then produces contextualized hidden states $\mathbf{H}=\mathcal{F}(\mathbf{X})\in\mathbb{R}^{L\times d}$, from which we extract the query-specific embeddings:
\begin{equation}
\mathbf{z}_k = \mathbf{H}[\mathbf{q}_k]\in\mathbb{R}^{d},\quad k=1,\ldots,K.
\end{equation}
Each $\mathbf{z}_k$ can be viewed as a grounded instance hypothesis: it integrates instruction semantics, global visual context, and query-level interactions, and is expected to encode both \emph{what} to segment, namely the semantic intent, and \emph{where} to segment, namely the localization cues. In this way, the query bank provides an explicit set-structured interface between instruction reasoning and downstream instance-level mask prediction.

\subsubsection{Hybrid-Attention Design}
To reconcile language generation with instance-level set prediction, we present a hybrid-attention pattern, as  illustrated in Fig.~\ref{fig: model}(b).
The key idea is to treat textual tokens and mask queries differently according to their roles, and instance queries should not be generated independently or sequentially.
Text tokens follow the standard causal attention used for autoregressive language modeling, while mask queries are allowed to attend bidirectionally to other mask queries. This design preserves the language modeling ability of the LLM, while enabling instance slots to communicate with each other to capture the target set structure and suppress duplicate predictions. 

Formally, let $\mathbf{A}\in\{0,1\}^{L\times L}$ be the attention mask. 
For text positions $i\in \mathcal{I}_{\text{text}}$, we enforce causal attention by setting $\mathbf{A}_{ij}=1$ only if $j\le i$. For query positions $i\in \mathcal{I}_{\text{query}}$, we allow full-context attention by setting $\mathbf{A}_{ij}=1$ for all $j\in(\mathcal{I}_{\text{vision}}\cup \mathcal{I}_{\text{text}}\cup \mathcal{I}_{\text{query}})$. In this way, each query obtains a global view of the image, the instruction, and the other instance slots, enabling more stable and instance-discriminative mask prediction.

\subsubsection{From Query to Mask}
To realize the reasoning-to-instance interface, we translate LLM-conditioned instance queries into detector-compatible prompts that directly control SAM3’s mask decoding process. 
Specially, 
a lightweight MLP projects each query embedding $\mathbf{z}_k$ into the embedding space expected by the mask decoder $\mathcal{D}$, yielding grounded mask-query embeddings $\{\mathbf{\tilde{z}_k\}_{k=1}^{K}}$. In parallel, another MLP maps the phrase features to the required dimensionality, producing $\mathbf{t}_p$ as auxiliary textual conditioning.
Given the projected features, the fusion encoder conditions  visual embeddings by cross-attending to the phrase tokens, producing instruction-aware image features. A subsequent detector then allows each mask query to cross-attend to these conditioned image features, refining instance-specific representations. Finally, a score head predicts the validity of each query, and a segmentation head generates its corresponding binary mask.

\subsection{Training Objectives}\label{subsec6}

We train InstructSAM end-to-end with a multi-task objective that jointly optimizes: (\textit{i}) a masked auto-regressive loss $\mathcal{L}_{\text{text}}$, (\textit{ii}) an instance segmentation loss $\mathcal{L}_{\text{seg}}$, and (\textit{iii}) a query-level presence loss $\mathcal{L}_{\text{pres}}$. The overall loss is
\begin{equation}
\label{eq:total_loss}
\mathcal{L}
=
\lambda_{\text{text}}\mathcal{L}_{\text{text}}
+
\lambda_{\text{seg}}\mathcal{L}_{\text{seg}}
+
\lambda_{\text{presence}}\mathcal{L}_{\text{presence}}.
\end{equation}

\noindent\textbf{Masked Auto-regressive Loss.}
Let $\mathbf{y}=(y_1,\ldots,y_N)$ denote the target text sequence produced by the MLLM, and let $\mathbf{x}$ denote the multimodal conditioning context, including instruction tokens and image tokens. We optimize the standard auto-regressive negative log-likelihood, while \emph{masking out} special segmentation-related tokens (e.g., instance query tokens $\mathbf{q}$ and \texttt{<mask\_end>}), so that they do not contribute to the language modeling objective. Specifically, we introduce a binary mask $m_i\in\{0,1\}$ indicating whether the $i$-th token is supervised by the text loss ($m_i=0$ for masked tokens). The masked auto-regressive loss is:
\begin{equation}
\label{eq:masked_ar_loss}
\mathcal{L}_{\text{text}}
=
-\frac{1}{\sum_{i=1}^{N} m_i}
\sum_{i=1}^{N}
m_i \log p_{\theta}\!\left(y_i \mid y_{<i}, \mathbf{x}\right).
\end{equation}

\noindent\textbf{Segmentation Loss.}
Following DETR-style set prediction~\cite{cheng2021mask2former,carion2020end}, we perform bipartite matching to compute an optimal one-to-one assignment between predicted instance slots and ground-truth instances. For matched slots, we supervise the predicted masks with a weighted combination of per-pixel binary cross-entropy and Dice loss:
\begin{equation}
\label{eq:seg_loss}
\mathcal{L}_{\text{seg}}
=
\lambda_{\text{bce}}\mathcal{L}_{\text{bce}}
+
\lambda_{\text{dice}}\mathcal{L}_{\text{dice}},
\end{equation}
where $\mathcal{L}_{\text{bce}}$ is computed over pixels and $\mathcal{L}_{\text{dice}}$ encourages overlap-aware mask quality. 

\noindent\textbf{Presence Loss.}
To identify which query slots correspond to valid target instances, we supervise each slot with a binary presence label. Specifically, after bipartite matching, we set $t_k=1$ if slot $k$ is matched to a ground-truth instance and $t_k=0$ otherwise.
We then apply a binary cross-entropy loss to the per-slot presence logits $\hat{s}_k$:
\begin{equation}
\label{eq:score_loss}
\mathcal{L}_{\text{presence}}
= \frac{1}{K}\sum_{k=1}^{K}
\mathrm{BCE}\!\left(\hat{s}_k, t_k\right).
\end{equation}

\section{\texttt{Inst$^{2}$Seg} Dataset}\label{sec4}

In this section, we present \textbf{\texttt{Inst$^{2}$Seg}}, a large-scale \textbf{inst}ruction-based \textbf{inst}ance \textbf{segmentation} dataset and benchmark designed to couple free-form instructions with instance-level masks. It is designed to support fine-grained instruction reasoning and precise mask annotation for complex instruction-driven segmentation.

\begin{table}[t]
  \centering
  \caption{Comparison of  \textbf{\texttt{Inst$^2$Seg}} benchmark with existing referring image segmentation benchmarks. Our \texttt{Inst$^{2}$Seg} provides instruction-based instance-level evaluation covering single-target, multi-target, no-target, and reasoning scenarios. 
  ST, MT, NT, and Reas. denote single-target, multi-targets, no-target, and reasoning, respectively.}
  \label{tab:comp}
  \setlength{\tabcolsep}{10pt}
  \renewcommand{\arraystretch}{1}
    \begin{tabular}{lccc|cc| c c}
      \toprule
      \textbf{Benchmarks} &
      \textbf{ST} &
      \textbf{MT} &
      \textbf{NT} &
      \makecell{\textbf{Inst.}\\\textbf{Level}} &
      \makecell{\textbf{Reas.}} &
      \makecell{\textbf{Prompt}\\\textbf{Type}} &
      \textbf{Metric} \\
      \midrule
      \midrule
      RefCOCO       & \cmark &        &        &        &        & Phrase       & IoU \\
      RefCOCO+      & \cmark &        &        &        &        & Phrase       & IoU \\
      RefCOCOg      & \cmark &        &        &        &        & Phrase       & IoU \\
      gRefCOCO      & \cmark & \cmark & \cmark & \cmark &        & Phrase       & IoU \\
      GSEval        & \cmark & \cmark &        &        &        & Phrase       & IoU \\
      \midrule
      ReasonSeg     & \cmark &        &        &        & \cmark & Instruction  & IoU \\
\rowcolor{blue!5}\textbf{\texttt{Inst$^{2}$Seg}} & \cmark & \cmark & \cmark & \cmark & \cmark & Instruction  & AP+IoU \\
      \bottomrule
    \end{tabular}
\end{table}

\noindent\textbf{Training Data.}
We collect training images from two sources: (\textit{i}) conventional exo-centric images sampled from SA-1B~\cite{kirillov2023segment}, COCO2017~\cite{lin2014microsoft}, and (\textit{ii}) ego-centric images curated from Ego4D~\cite{grauman2022ego4d}, EPIC-KITCHENS~\cite{Damen2018EPICKITCHENS}, and HD-EPIC~\cite{perrett2025hd}. For ego-centric subset, we crop clips with substantial scene variation and discard blurry or low-quality frames. Our annotation pipeline consists of four stages. (1) \textbf{QA generation} using Gemini 3 Flash~\cite{team2023gemini} to produce localization-oriented referring questions with hard negatives and concise noun-phrase answers, along with an explicit ground field encoding counting/quantifiers for multi-instance targets; (2) \textbf{object consolidation \& box generation}, where questions referring to the same target are merged into a shared \texttt{object\_id} and Gemini predicts normalized 2D boxes; (3) \textbf{mask annotation} by prompting SAM2~\cite{ravi2024sam} with the boxes to obtain pixel-accurate instance masks per \texttt{object\_id}; and (4) \textbf{filtering} to remove low-quality or inconsistent samples. In total, we curate 100K images with 500K QA pairs.

\begin{figure}[t]
\centering
\includegraphics[width=0.70\linewidth]{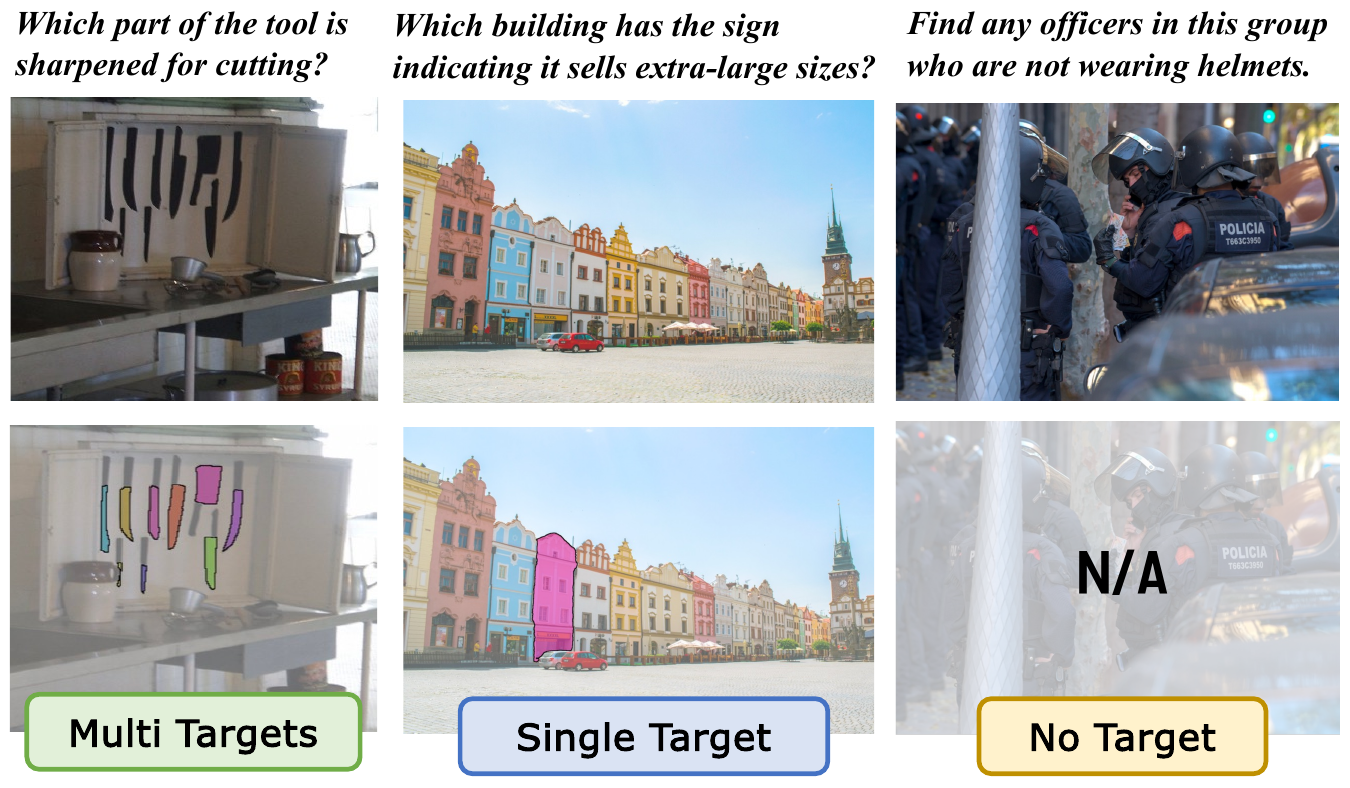}
\caption{Examples of the proposed \textbf{\texttt{Inst$^2$Seg}} benchmark, covering diverse instruction types that require instance-level reasoning and segmentation.}\label{fig:benchmark}
\end{figure}

\noindent\textbf{Benchmark.}
The \textbf{\texttt{Inst$^{2}$Seg}} benchmark comprises 986 images and 3,328 unique instructions. Compared with existing referring image segmentation benchmarks~\cite{kazemzadeh2014referitgame,mao2016generation,he2023grec,hu2025groundingsuite} (Table~\ref{tab:comp}), \textbf{\texttt{Inst$^{2}$Seg}} provides a more challenging evaluation setting by covering single-target, multi-target, and no-target cases under instruction prompts, spanning both object-level and part-level granularity. All benchmark instructions and masks are manually verified to ensure high quality. Representative examples are shown in Fig.~\ref{fig:benchmark}.

\noindent\textbf{Metrics.}
We adopt mean Average Precision (mAP) as the primary metric for evaluating instance-level predictions. To provide a more fine-grained analysis, we further stratify the results by the number of targets, including single-target, multi-target, and no-target cases. Since many prior methods do not explicitly distinguish individual instances, we additionally report generalized IoU (gIoU) as a complementary semantic-level metric.

\section{Experiments}\label{sec5}

\subsection{Implementation Details}\label{subsec7}

Our framework is built on the Qwen3-VL~\cite{Qwen3-VL} backbone, with the mask decoder initialized from SAM 3~\cite{li2025sam3}.  We set the number of learnable queries $K=10$ by default. For parameter-efficient fine-tuning, we apply LoRA to the LLM with a rank of 256. 
Training is conducted in two stages: the first stage aligns the learnable query space with SAM3 for referring segmentation, while the second stage injects reasoning-oriented instruction knowledge. Training is conducted in two stages with different data compositions.
The first stage 
focuses on alignment for referring segmentation, while the second stage aims to inject reasoning-oriented instruction knowledge. Detailed training data for each stage are provided in Appendix~\ref{secA1}.
Each stage is trained for one epoch.
For optimization, we set 
$\lambda_{\text{text}}=\lambda_{\text{mask}}=\lambda_{\text{score}}=1.0$. In the segmentation loss, the weights for BCE and Dice losses are set to $\lambda_{\text{bce}}=2.0$ and $\lambda_{\text{dice}}=0.5$, respectively.

\begin{table}[t]
\caption{Results on \textbf{reasoning-based instruction segmentation benchmarks}, including our proposed \textbf{\texttt{Inst$^{2}$Seg}} (instance-level) and ReasonSeg~\cite{lai2024lisa} (semantic-level). \textcolor{gray}{Grey}  entries indicate that  models are finetuned on the corresponding training data.}\label{tab:reasoning}
\centering
\setlength{\tabcolsep}{2.2pt}
\renewcommand{\arraystretch}{1.1}
\footnotesize
\begin{tabular}{l *{7}{c} *{8}{c}}
\toprule
\multirow{2}{*}{\textbf{Method}}
  & \multicolumn{7}{c}{\textbf{\texttt{Inst$^{2}$Seg}}}
  & \multicolumn{8}{c}{\textbf{ReasonSeg}} \\
\cmidrule(lr){2-8}\cmidrule(lr){9-16}
  & \multicolumn{2}{c}{\textbf{Overall}}
  & \multicolumn{2}{c}{\textbf{Single-Target}}
  & \multicolumn{2}{c}{\textbf{Multi-Target}}
  & \multicolumn{1}{c}{\textbf{No-Target}}
  & \multicolumn{2}{c}{\textbf{val}}
  & \multicolumn{2}{c}{\textbf{test (all)}}
  & \multicolumn{2}{c}{\textbf{test (short)}}
  & \multicolumn{2}{c}{\textbf{test (long)}} \\
\cmidrule(lr){2-3}\cmidrule(lr){4-5}\cmidrule(lr){6-7}\cmidrule(lr){8-8}
\cmidrule(lr){9-10}\cmidrule(lr){11-12}\cmidrule(lr){13-14}\cmidrule(lr){15-16}
  & {mAP} & {gIoU}
  & {mAP} & {gIoU}
  & {mAP} & {gIoU}
  & {gIoU}
  & {gIoU} & {cIoU}
  & {gIoU} & {cIoU}
  & {gIoU} & {cIoU}
  & {gIoU} & {cIoU} \\
\midrule
\rowcolor{black!5}
\multicolumn{16}{c}{\textit{Multi-round Agent Pipeline}} \\
\midrule
SAM3-Agent$_{\text{Qwen2.5-VL-3B}}$~\cite{carion2025sam} & 23.2 & 48.7 & 33.9 & 47.6 & 18.8 & 33.1 & 86.2 & 50.3 & 34.1 & 49.9 & 46.4 & 44.8 & 36.2 & 50.0 & 47.5 \\
SAM3-Agent$_{\text{Qwen2.5-VL-7B}}$~\cite{carion2025sam}
  & \textbf{35.7} & \textbf{63.2} & \textbf{47.8} & \textbf{65.1} & \textbf{30.4} & \textbf{47.9} & \textbf{90.1}
  & \textbf{62.2} & \textbf{49.1}
  & \textbf{63.0} & \textbf{53.5}
  & \textbf{59.4} & \textbf{43.5}
  & \textbf{64.1} & \textbf{56.2} \\
SAM3-Agent$_{\text{Qwen3-VL-2B}}$~\cite{carion2025sam} & 29.7 & 58.8 & 43.2 & 62.1 & 24.4 & 43.2 & 81.3 & 58.0 & 44.9 & 56.5 & 43.6 & 54.1 & 36.1 & 57.3 & 45.9  \\

\midrule
\rowcolor{black!5}
\multicolumn{16}{c}{\textit{End-to-end Model}} \\
\midrule
LISA-7B~\cite{lai2024lisa}
  & 1.9 & 28.6 & 8.2 & 37.7 & 0.2 & 27.7 & 0.2
  & \textcolor{gray}{52.9} & \textcolor{gray}{54.0}
  & \textcolor{gray}{47.3} & \textcolor{gray}{48.4}
  & \textcolor{gray}{40.6} & \textcolor{gray}{40.6}
  & \textcolor{gray}{49.4} & \textcolor{gray}{51.0} \\
LISA++-7B~\cite{yang2023lisa++}
  & 2.2 & 25.8 & 7.7 & 34.1 & 0.5 & 24.9 & 0.2
  & \textcolor{gray}{\underline{64.2}} & \textcolor{gray}{\textbf{68.1}}
  & \textcolor{gray}{57.0} & \textcolor{gray}{\underline{59.5}}
  & \textcolor{gray}{49.6} & \textcolor{gray}{\underline{51.1}}
  & \textcolor{gray}{59.3} & \textcolor{gray}{\underline{61.7}} \\
PixelLM-7B~\cite{ren2024pixellm}
  & 4.6 & 27.2 & 13.9 & 36.7 & 0.8 & 22.9 & 5.1
  & 44.7 & 37.4 & 44.0 & 39.7 & 38.4 & 36.7 & 45.8 & 40.6 \\
SA2VA-4B~\cite{yuan2025sa2va}
  & 8.2 & 52.6 & 30.1 & 59.2 & 1.1 & 40.2 & \underline{56.9}
  & 59.2 & 60.0 & 56.9 & 55.8 & 53.1 & \textbf{53.0} & 58.2 & 56.4 \\
SA2VA-8B~\cite{yuan2025sa2va}
  & 9.4 & \underline{53.9} & \underline{35.5} & \underline{65.2} & 1.4 & \textbf{46.3} & 32.4
  & \textbf{65.2} & 61.1 & \underline{59.5} & 57.8 & \underline{55.1} & 49.8 & \underline{60.9} & 60.0 \\
X-SAM-3.8B~\cite{wang2025x}
  & \underline{11.0} & 36.6 & 33.4 & 51.9 & \underline{2.0} & 30.0 & 0.0
  & 56.6 & 32.9 & 57.8 & 41.0 & 47.7 & 48.1 & 56.0 & 40.8 \\
\rowcolor{blue!5}\textbf{InstructSAM-2B}
  & \textbf{31.5} & \textbf{60.4} & \textbf{52.6} & \textbf{66.8} & \textbf{22.2} & \underline{44.0} & \textbf{74.3}
  & 62.5 & \underline{65.0} & \textbf{61.1} & \textbf{61.0} & \textbf{56.0} & \underline{51.1} & \textbf{62.7} & \textbf{63.3} \\
\bottomrule
\end{tabular}
\end{table}

\subsection{Complex Reasoning Segmentation}\label{subsec8}

\noindent\textbf{Instance-level Inst$^{2}$Seg.}
Table~\ref{tab:reasoning} reports results on the Inst$^{2}$Seg benchmark across three subsets: \emph{single-target}, \emph{multi-target}, and \emph{no-target}.
For multi-round agentic pipeline, we evaluate SAM3-Agent~\cite{carion2025sam} with Qwen2.5-VL-3B and Qwen2.5-VL-7B. For end-to-end models, we compare with semantic-level methods, including  LISA~\cite{lai2024lisa}, SA2VA~\cite{yuan2025sa2va}, and X-SAM~\cite{wang2025x}, 
as well as instance-level methods LISA++~\cite{yang2023lisa++} and PixelLM~\cite{ren2024pixellm}.

Among end-to-end methods, InstructSAM-2B achieves the best mAP by a large margin, 
highlighting the  superiority of instance-level instruction following. 
To complement mAP, we report gIoU as a semantic metric by taking the union of predicted masks per sample; the gap between gIoU and mAP indicates that instance discrimination is substantially harder than coarse semantic region localization, yet our InstructSAM remains strong on most subsets. 
Notably, for the no-target subset, we evaluate in a zero-shot setting without using any no-target instructions from the Inst$^{2}$Seg training set. InstructSAM still achieves robust performance, demonstrating its generalization to invalid-target cases.
Compared with multi-round agentic pipelines that require multiple interaction rounds and longer contexts, InstructSAM delivers leading performance at comparable model scale. For example, compared with SAM3-Agent based on Qwen2.5-VL-3B, InstructSAM improves mAP by +8.3 and gIoU by +11.7.
These results indicate the effectiveness of InstructSAM 
for instruction-driven segmentation.

\noindent\textbf{Semantic-level ReasonSeg.}
Table~\ref{tab:reasoning} also reports results on ReasonSeg benchmark for the reasoning semantic segmentation. Following the official protocol of~\cite{lai2024lisa}, we report gIoU and cIoU on both validation and test splits. Compared to similarly sized models such as X-SAM~\cite{wang2025x} and SA2VA-4B~\cite{yuan2025sa2va}, InstructSAM achieves substantially better performance, improving cIoU by +5.0 on the validation split and +5.2 on the overall test set. 
The improvement is especially notable on long instructions, where InstructSAM gains +6.9 cIoU on \textit{test (long)}, demonstrating stronger robustness to complex and lengthy descriptions.

\begin{table}[t]
\caption{Results on \textbf{phrase-level Referring Expression Segmentation benchmarks}, including multi-object gRefCOCO and zero-shot GSEval. We report instance-level mAP and semantic-level cIoU/gIoU where applicable.
}\label{tab:phrase}
\centering
\setlength{\tabcolsep}{8.0pt}
\renewcommand{\arraystretch}{1.05}
\footnotesize
\begin{tabular}{l c c | c c | c c | c c c c c}
\toprule
\multirow{3}{*}{\textbf{Methods}}
  & \multicolumn{6}{c}{\textbf{gRefCOCO}}
  & \multicolumn{5}{c}{\textbf{GSEval}} \\
\cmidrule(lr){2-7}\cmidrule(lr){8-12}
  & \multicolumn{2}{c}{\textbf{val}}
  & \multicolumn{2}{c}{\textbf{testA}}
  & \multicolumn{2}{c}{\textbf{testB}}
  & \textbf{Stuff} & \textbf{Part} & \textbf{Multi} & \textbf{Single} & \textbf{All} \\
\cmidrule(lr){2-3}\cmidrule(lr){4-5}\cmidrule(lr){6-7}\cmidrule(lr){8-12}
  & {mAP} & {cIoU}
  & {mAP} & {cIoU}
  & {mAP} & {cIoU}
  & {gIoU} & {gIoU} & {gIoU} & {gIoU} & {gIoU} \\
\midrule
LISA-7B~\cite{lai2024lisa}     & 28.1 & 53.9 & 41.5 & 63.6 & 31.9 & 55.6 & 85.2 & 21.2 & 71.5 & 42.8 & 57.6 \\
GLAMM~\cite{rasheed2024glamm}  & \na & \na & \na & \na & \na & \na & 86.9 & 16.5 & 70.4 & 42.1 & 57.2 \\
GSVA-7B~\cite{xia2024gsva}     & \na & 61.7 & \na & 69.2 & \na & 60.3 & 76.0 & 20.0 & 57.8 & 34.2 & 48.6 \\
PSALM~\cite{zhang2024psalm}    & \na & 42.0 & \na & 52.4 & \na & 50.6 & 39.0 & 10.0 & 53.7 & 36.9 & 37.7 \\
EVF-SAM~\cite{zhang2024evf}    & \na & \na & \na & \na & \na & \na & 85.1 & 23.1 & 72.1 & 54.5 & 62.6 \\
InstructSeg~\cite{wei2025instructseg} & \na & \na & \na & \na & \na & \na & 56.2 & \textbf{24.2} & 66.8 & 51.3 & 52.5 \\
PixelLM-7B~\cite{ren2024pixellm} & 33.9 & 51.2 & 30.7 & 62.6 & 23.9 & 54.6 & \na & \na & \na & \na & \na \\
SA2VA-4B~\cite{yuan2025sa2va}  & 26.1 & 42.3 & 36.6 & 53.4 & 32.6 & 47.9 & 88.5 & 17.4 & 68.4 & 45.5 & 58.5 \\
SA2VA-8B~\cite{yuan2025sa2va}  & 25.2 & 42.9 & 37.7 & 55.4 & 32.1 & 48.9 & 77.3 & 18.0 & 72.0 & 45.5 & 56.3 \\
X-SAM-3.8B~\cite{wang2025x}         & 22.6 & 37.0 & 31.7 & 46.6 & 30.4 & 42.9 & \na & \na & \na & \na & \na \\
\midrule
\rowcolor{blue!5}\textbf{InstructSAM-2B} & \textbf{57.3} & \textbf{68.3} & \textbf{51.9} & \textbf{72.3} & \textbf{43.5} & \textbf{65.2} & \textbf{89.4} & 22.4 & \textbf{73.6} & \textbf{54.8} & \textbf{64.1} \\
\bottomrule
\end{tabular}
\end{table}

\subsection{Phrase-level Referring Expression Segmentation}\label{subsec9}

\noindent\textbf{gRefCOCO.}
As shown in Table~\ref{tab:phrase}, InstructSAM achieves strong performance on the gRefCOCO~\cite{he2023grec} benchmark. Since gRefCOCO provides instance-level annotations and contains multiple target instances, we report instance-level mAP in addition to the conventional IoU-based metric. InstructSAM-2B surpasses the strongest prior method, GSVA-7B~\cite{xia2024gsva}, in semantic-level cIoU with gains of +6.6 on val, +3.1 on testA, and +4.9 on testB. It also achieves substantially higher mAP than multi-instance-capable methods such as PixelLM-7B~\cite{ren2024pixellm} and X-SAM~\cite{wang2025x}, demonstrating robust instance discrimination across diverse referring expressions.

\noindent\textbf{GSEval.}
Table~\ref{tab:phrase} reports zero-shot results on GSEval~\cite{hu2025groundingsuite}, 
a comprehensive referring expression segmentation benchmark covering four challenging subsets: \emph{stuff}, \emph{part}, \emph{multi-object}, and \emph{single-object}. Since GSEval provides only semantic-level annotations, we follow the official protocol and report gIoU. InstructSAM achieves the best performance, outperforming the previous state-of-the-art method, EVF-SAM~\cite{zhang2024evf}, by +1.5 gIoU.

\begin{wraptable}[12]{r}{0.5\textwidth}
\vspace{-3mm}
\caption{Results on RoboRefIt benchmark. InstructSAM-2B outperforms previous segmentation methods and shows strong generalization on the distribution-shifted testB split.}\label{tab: roborefit}
\vspace{-2mm}
\centering
\setlength{\tabcolsep}{10pt}
\footnotesize
\resizebox{\linewidth}{!}{%
\begin{tabular}{l c c c}
\toprule
\textbf{Method} & \textbf{Task} & \multicolumn{2}{c}{\textbf{RoboRefIt}} \\
\cmidrule(lr){3-4}
 & \textbf{Specific} & \textbf{testA} & \textbf{testB} \\
\midrule
RefTR-r50~\cite{lu2023vl} & \cmark & \textbf{85.5} & 61.5 \\
RefTR-r101~\cite{lu2023vl} & \cmark & 83.9 & 60.7 \\
\midrule
LISA-7B~\cite{lai2024lisa}  & \xmark & 36.1 & 28.7  \\
SA2VA-4B~\cite{yuan2025sa2va}  & \xmark & 56.8 & 34.0 \\
\rowcolor{blue!5}\textbf{InstructSAM-2B} & \xmark & 82.5 & \textbf{74.4} \\
\bottomrule
\end{tabular}%
}
\vspace{-3mm}
\end{wraptable}
\noindent\textbf{RoboRefIt.}
We further evaluate InstructSAM on RoboRefIt~\cite{lu2023vl}, a challenging visual grounding benchmark designed for robotic perception and reasoning in indoor environments. RoboRefIt requires a robot to localize the target object specified by language commands. As shown in Table~\ref{tab: roborefit}, InstructSAM-2B substantially outperforms other MLLM-based segmentation methods. Notably, it approaches the task-specific RefTR-r50~\cite{lu2023vl} on the in-distribution testA split and exceeds it by a large margin on the distribution-shifted testB split (+12.9), highlighting the strong generalization ability of our approach.

\subsection{Ablation Studies}\label{subsec10}

\noindent\textbf{Key Designs in InstructSAM.} We ablate two core components of InstructSAM in Table~\ref{tab: ablation_module}. 
\begin{wraptable}{r}{0.55\textwidth}
\vspace{-3mm}
\caption{Ablation study of key design choices in InstructSAM. We report the average validation score on gRefCOCO, mAP on Inst$^2$Seg, and cIoU on ReasonSeg-val.}\label{tab: ablation_module}
\centering
\setlength{\tabcolsep}{2.6pt}
\begin{tabular}{lccc}
\toprule
& \textbf{gRefCOCO} & \textbf{Inst$^2$Seg} & \textbf{ReasonSeg} \\
\midrule
\textit{w/o} Query Bank   & 58.3 & 20.1 & 56.0 \\
\textit{w/o} Hybrid-Attention     & 61.2 & 29.5 & 52.4 \\
\rowcolor{blue!5}\textbf{InstructSAM}   & \textbf{62.8} & \textbf{31.5} & \textbf{65.0} \\
\bottomrule
\end{tabular}%
\vspace{-3mm}
\end{wraptable}
Removing the learnable queries forces the model to rely on autoregressively generated mask tokens without explicit query conditioning, resulting in consistent performance drops across all benchmarks. 
The degradation is most pronounced on the instance-level \texttt{Inst$^{2}$Seg} benchmark, where mAP decreases from 31.5 to 20.1, demonstrating the importance of query bank for instance grounding.
Replacing hybrid attention with plain causal attention also degrades performance, especially on ReasonSeg, where the score drops from 65.0 to 52.4 Replacing hybrid-attention with plain causal attention (\textit{w/o hybrid-attention}) also degrades performance, especially on ReasonSeg, where the score drops from 65.0 to 52.4. This highlights the role of bidirectional query interaction and cross-modal fusion in enabling robust visual-text grounding and mask prediction.

\begin{table}[t]
\caption{
Effect of query number on inference efficiency and segmentation performance. Larger query banks bring marginal performance changes but higher latency;
``s'' and ``m'' denote single-target and multi-target settings, respectively.}\label{tab: supp_query_scaling}
\centering
\small
\setlength{\tabcolsep}{10pt}
\resizebox{0.8\linewidth}{!}{%
\begin{tabular}{lccccc}
\toprule
\textbf{Query Num} & \textbf{Infer Time (s)} $\downarrow$ & \makecell{\textbf{Inst\textsuperscript{2}Seg-s} \\ \textbf{(mAP)}} & \makecell{\textbf{Inst\textsuperscript{2}Seg-m} \\ \textbf{(mAP)}} & \makecell{\textbf{Inst\textsuperscript{2}Seg} \\ \textbf{(mAP)}} & \makecell{\textbf{ReasonSeg} \\ \textbf{(gIoU)}} \\
\midrule
\midrule
10 & \textbf{1.1} & \textbf{52.6} & 22.2 & \textbf{31.5} & 62.5 \\
50 & 1.4 & 52.1 & \textbf{22.3} & 31.3 & \textbf{62.9} \\
200 & 2.1 & 52.0 & 22.0 & 31.1 & 62.3 \\
\bottomrule
\end{tabular}
}
\vspace{-2mm}
\end{table}

\begin{table}[t]
\caption{Ablation study on phrase conditioning and LLM-conditioned queries. The results show that LLM-conditioned queries are the primary carrier of instruction semantics, while the generated phrase mainly serves as auxiliary conditioning.}\label{tab: supp_llm_conditioned_query}
\centering
\small
\setlength{\tabcolsep}{8pt}
\resizebox{0.8\linewidth}{!}{%
\begin{tabular}{lcccc}
\toprule
\textbf{Method} & \makecell{\textbf{Inst\textsuperscript{2}Seg} \\ \textbf{mAP}} & \makecell{\textbf{Inst\textsuperscript{2}Seg} \\ \textbf{gIoU}} & \makecell{\textbf{ReasonSeg val} \\ \textbf{gIoU}} & \makecell{\textbf{ReasonSeg test} \\ \textbf{gIoU}} \\
\midrule
\rowcolor{blue!5}
\textbf{Ours} & \textbf{31.5} & \textbf{60.4} & \textbf{62.5} & \textbf{61.1} \\
w/o phrase (dummy token) & 29.1 (\drop{2.4}) & 58.5 (\drop{1.9}) & 60.7 (\drop{1.8}) & 58.9 (\drop{2.2}) \\
w/o LLM-conditioned query & 16.7 (\drop{14.8}) & 42.1 (\drop{18.3}) & 45.0 (\drop{17.5}) & 42.9 (\drop{18.2}) \\
\bottomrule
\end{tabular}
}
\end{table}

\begin{table}[t]
\centering
\begin{minipage}[t]{0.53\linewidth}
\centering
\caption{Effectiveness of the data engine, including the impact of the Inst$^2$Seg data and the filtering strategy.}\label{tab: ablation_data}
\setlength{\tabcolsep}{8pt}
\scriptsize
\resizebox{\linewidth}{!}{%
\begin{tabular}{lccccc}
\toprule
\multirow{2}{*}{\textbf{Setting}} &
\multicolumn{3}{c}{\textbf{Inst$^2$Seg}} &
\multicolumn{2}{c}{\textbf{ReasonSeg}} \\
\cmidrule(lr){2-4}\cmidrule(lr){5-6}
& \textbf{ST} & \textbf{MT} & \textbf{All} & \textbf{cIoU} & \textbf{gIoU} \\
\midrule
\textit{w/o} Inst$^2$Seg Data          & 47.6 & 17.6 & 25.9 & 62.1 & 60.3 \\
\textit{w/o} Filtering & 10.2 & 13.1 & 11.9 & 57.9 & 58.1  \\
\rowcolor{blue!5} Full Data Engine       & \textbf{52.6} & \textbf{22.2} & \textbf{31.5} & \textbf{63.0} & \textbf{61.8}\\
\bottomrule
\end{tabular}%
}
\end{minipage}
\hfill
\begin{minipage}[t]{0.44\linewidth}
\centering
\caption{Ablation  on the referring alignment pretraining (stage 1). This alignment stage substantially improves both referring and instruction-based segmentation performance.}\label{tab:ablation_stage}
\setlength{\tabcolsep}{4pt}
\scriptsize
\resizebox{\linewidth}{!}{%
\begin{tabular}{ccccc}
\toprule
\textbf{Stage 1} &
\makecell{\textbf{gRefCOCO}\\\textbf{val(mAP)}} &
\makecell{\textbf{gRefCOCO}\\\textbf{val(cIoU)}} &
\makecell{\textbf{Inst\textsuperscript{2}Seg}\\\textbf{(mAP)}} &
\makecell{\textbf{ReasonSeg}\\\textbf{val(cIoU)}} \\
\midrule
\rowcolor{blue!5}\cmark & \textbf{57.3} & \textbf{68.3} & \textbf{31.5} & \textbf{65.0} \\
\xmark & 41.3\,(\drop{16.0}) & 57.7\,(\drop{10.6}) & 8.1\,(\drop{23.4}) & 15.9\,(\drop{49.1}) \\
\bottomrule
\end{tabular}%
}
\end{minipage}
\end{table}

\noindent\textbf{Scaling with Query Number.}
We analyze the sensitivity of InstructSAM to the query number $K$ and scalability to dense scenes. As shown in Table~\ref{tab: supp_query_scaling}, increasing $K$ from 10 to 200 leads to minimal performance change while inference time steadily increases. This suggests that a small number of well-conditioned queries is sufficient for most samples, consistent with the data distribution where the majority of instructions involve fewer than 10 target instances. Larger query banks may introduce redundant slots without clear performance gains.

\noindent\textbf{Significance of LLM-conditioned query.}
To verify that the primary semantic signal comes from the LLM-conditioned query representations, we conduct two inference ablations using the same trained checkpoint. First, we replace the generated noun phrase with a \textit{dummy token} (token id 0) while keeping the rest of the architecture unchanged. This  causes only a modest performance drop, suggesting that the noun phrase is not the main carrier of instruction semantics. Second, we remove the LLM-conditioned query representations. As shown in Table~\ref{tab: supp_llm_conditioned_query}, this leads to much larger degradation across all metrics, demonstrating that the dominant semantic information is encoded in the instruction-conditioned queries after decoder fusion.
These results indicate that the noun phrase mainly provides auxiliary textual conditioning for stability and compatibility with the SAM3 interface, while the LLM-conditioned queries serve as the core semantic representation for instruction-driven mask prediction.

\begin{figure*}[t]
\centering
\includegraphics[width=0.99\linewidth]{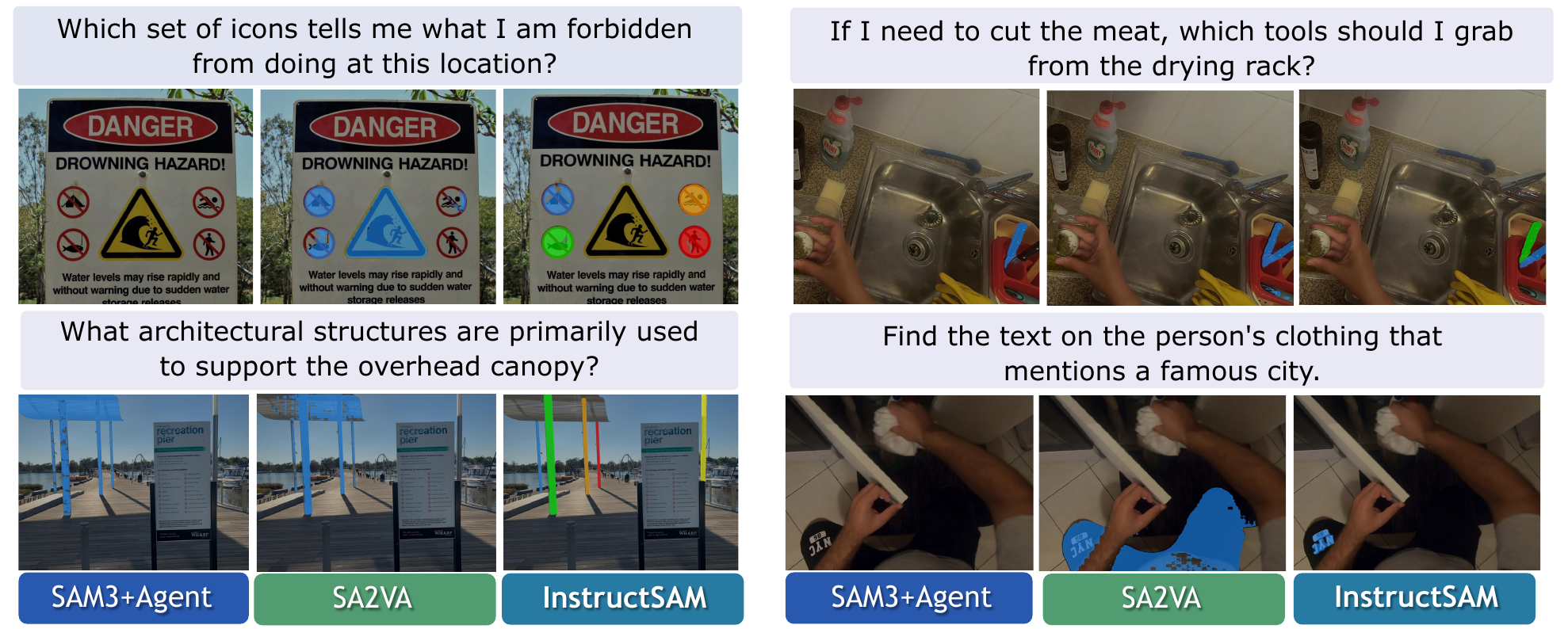}
\caption{\textbf{Qualitative comparison with SAM3-Agent-Qwen3-VL-2B~\cite{Qwen3-VL} and SA2VA-4B~\cite{yuan2025sa2va}}. InstructSAM better understands complex instructions and produces more accurate instance-level masks, especially in scenarios requiring fine-grained reasoning, object distinction, and multi-target localization. More qualitative results are provided in Appendix~\ref{secA3}.}\label{fig:comp}
\end{figure*}

\noindent\textbf{Effectiveness of the Data Engine.}
Table~\ref{tab: ablation_data} ablates our data engine on both the instance-level benchmark \texttt{Inst$^{2}$Seg} and the semantic-level benchmark ReasonSeg. Removing the constructed training data (\textit{w/o training data}) reduces performance on \texttt{Inst$^{2}$Seg}, showing that high-quality multi-instance instruction-mask pairs are crucial for learning reliable instance-aware behavior. 
More importantly, removing the filtering stage (\textit{w/o filtering}) causes substantial degradation on both benchmarks, with \texttt{Inst$^{2}$Seg} mAP dropping from 31.5 to 11.9 and ReasonSeg cIoU/gIoU decreasing from 63.0/61.8 to 57.9/58.1. This suggests that naive MLLM--SAM3 data generation introduces considerable label noise. The full pipeline achieves the best performance, indicating that our data engine provides reliable supervision for instruction-driven instance segmentation.

\noindent\textbf{Effect of the Alignment Stage.}
Table~\ref{tab:ablation_stage} evaluates the alignment stage (stage 1) on referring and instruction-based segmentation benchmarks. Removing alignment causes consistent drops: on gRefCOCO, val mAP decreases from 57.3 to 41.3 (-16.0) and val cIoU drops from 68.3 to 57.7 (-10.6). The degradation is even more severe on instruction-following benchmarks, where \texttt{Inst$^{2}$Seg} mAP drops from 31.5 to 8.1 (-23.4) and ReasonSeg val cIoU drops from 65.0 to 15.9 (-49.1). These results show that the alignment stage is critical for coupling language understanding with mask prediction, and improving generalization to compositional instructions.

\begin{wraptable}[5]{r}{0.38\textwidth}
\vspace{-4.8mm}
\caption{Inference-time comparison on \texttt{Inst\textsuperscript{2}Seg}  benchmark.}\label{tab: supp_infer_time}
\vspace{-2.5mm}
\centering
\scriptsize
\resizebox{\linewidth}{!}{%
\begin{tabular}{lc}
\toprule
\textbf{Model} & \textbf{Infer Time (s)} $\downarrow$ \\
\midrule
\rowcolor{blue!5}InstructSAM-2B & \textbf{1.1} \\
SAM3-Agent-Qwen3-VL-2B & 29.6 \\
\bottomrule
\end{tabular}%
}
\end{wraptable}

\noindent\textbf{Inference Efficiency.}
We  compare inference efficiency under a controlled setting. All methods use the same Qwen3-VL-2B backbone and are evaluated under identical hardware conditions. Latency is measured on \texttt{Inst\textsuperscript{2}Seg} and averaged over all instructions.  As reported in Table~\ref{tab: supp_infer_time}, InstructSAM is substantially faster than SAM3-Agent, highlighting the efficiency advantage of our unified framework over multi-step agentic execution.

\subsection{Qualitive Results}
Fig.~\ref{fig:comp} presents qualitative comparisons between InstructSAM and existing methods, including SAM3-Agent-Qwen3-VL-2B~\cite{Qwen3-VL} and SA2VA-4B~\cite{yuan2025sa2va}. The examples cover diverse instruction-following scenarios, such as identifying forbidden signs, selecting tools, recognizing supporting structures, and locating text on clothing. Compared with the baselines, InstructSAM produces more accurate instance-level masks and better follows complex language instructions. In particular, it shows stronger capability in fine-grained visual reasoning and multi-target localization.

\section{Conclusion}\label{sec6}

In this paper, we presented InstructSAM, a unified framework for instruction-driven multi-instance segmentation with arbitrary complex instructions.
Built upon an explicit reasoning-to-instance query interface, InstructSAM translates arbitrary instructions into a set of parallel, learnable instance slots and projects them into SAM3's detector query space.
This design enables robust reasoning over free-form instructions involving attributes, relations, counting, exclusion and implicit intent.
Across diverse instruction-driven and referring segmentation benchmarks, the compact 2B-scale InstructSAM consistently achieves strong performance, producing accurate and efficient multi-instance masks while substantially outperforming prior end-to-end methods and even surpassing multi-round agentic pipeline.

\newpage
\appendix

\renewcommand{\thefigure}{\Alph{section}.\arabic{figure}}
\renewcommand{\thetable}{\Alph{section}.\arabic{table}}

\counterwithin{figure}{section}
\counterwithin{table}{section}

\section*{Appendix}\label{appendix}

\section{More Training Details}\label{secA1}
\begin{figure*}[h]
\centering
\includegraphics[width=\linewidth]{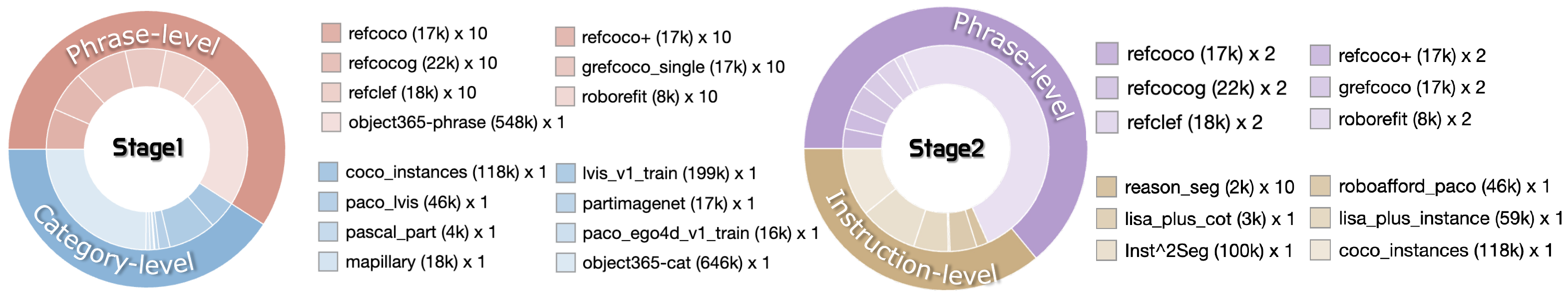}
\caption{The distribution of training datasets for InstructSAM.}\label{fig:dataset}
\end{figure*}

\noindent\textbf{Stage 1: Referring Alignment Pretraining.}
In this stage, we pretrain the learnable mask queries and explicitly align the mask-query space produced by the LLM with that of SAM3. To achieve robust alignment, we leverage a large collection of category-level and phrase-level simple mask grounding data, as illustrated in Fig.~\ref{fig:dataset} (left). In total, 2.5M training samples are used for alignment pretraining. We optimize all parameters except the vision encoder in the mask decoder, which is kept frozen. The global batch size is set to 128. We use a learning rate of $5\times 10^{-6}$ for the MLLM, including the LLM, vision encoder, and projector, and $1\times 10^{-5}$ for the remaining trainable modules in the mask decoder. The MLLM is initialized from Qwen3-VL-2B, while the mask decoder is initialized from the pretrained SAM3.

\noindent\textbf{Stage 2: Reasoning Knowledge Fine-tuning.}
Starting from the aligned model obtained in Stage~1, we further fine-tune the system to incorporate reasoning-aware segmentation knowledge and enhance instruction-following for more complex referring and compositional scenarios. As shown in Fig.~\ref{fig:dataset} (right), Stage~2 combines instruction-level segmentation data with a subset of phrase-level referring data, maintaining grounding robustness while improving reasoning generalization. The total number of training samples in this stage is 0.5M. We continue to freeze the vision encoder of the mask decoder and fine-tune the remaining components with a smaller batch size of 64 for stable adaptation. The learning rates are reduced to $2\times 10^{-6}$ for the LLM, the MLLM vision encoder, and the projector, and set to $5\times 10^{-6}$ for other trainable modules. This stage is also trained for one epoch.

The detailed hyper-parameters for the multi-stage training are reported in Table~\ref{tab: stage}, and the dataset distribution is provided in Fig.~\ref{fig:dataset}.

\begin{table}[h]
\caption{The Hyper-parameters in multi-stage training of InstructSAM. VE denotes vision encoder.}\label{tab: stage}
\centering
\setlength{\tabcolsep}{18pt}
\renewcommand{\arraystretch}{0.98}
\begin{tabular}{lcc}
\toprule
\textbf{Item} & \textbf{Stage1} & \textbf{Stage2} \\
\midrule
batch size            & 128    & 64  \\
training epochs       & 1    & 1    \\
lr of LLM   & 5e-6    & 2e-6   \\
lr of VE in MLLM  & 5e-6  & 2e-6   \\
lr of projector       & 5e-6  & 2e-6   \\
lr of VE in mask decoder  & 0  & 0  \\
lr of other modules   & 1e-5  & 5e-6 \\
optimizer             & \multicolumn{2}{c}{AdamW} \\
optimizer momentum    & \multicolumn{2}{c}{$\beta_1=0.9,\ \beta_2=0.999$} \\
weight decay          & 0.0  & 0.0    \\
warmup ratio          & 0.03  & 0.03 \\
LoRA rank & 64 & 64 \\
\bottomrule
\end{tabular}
\vspace{-2mm}
\end{table}

\section{More Visualization Results}\label{secA3}

\begin{figure*}[h!]
\centering
\includegraphics[width=\linewidth]{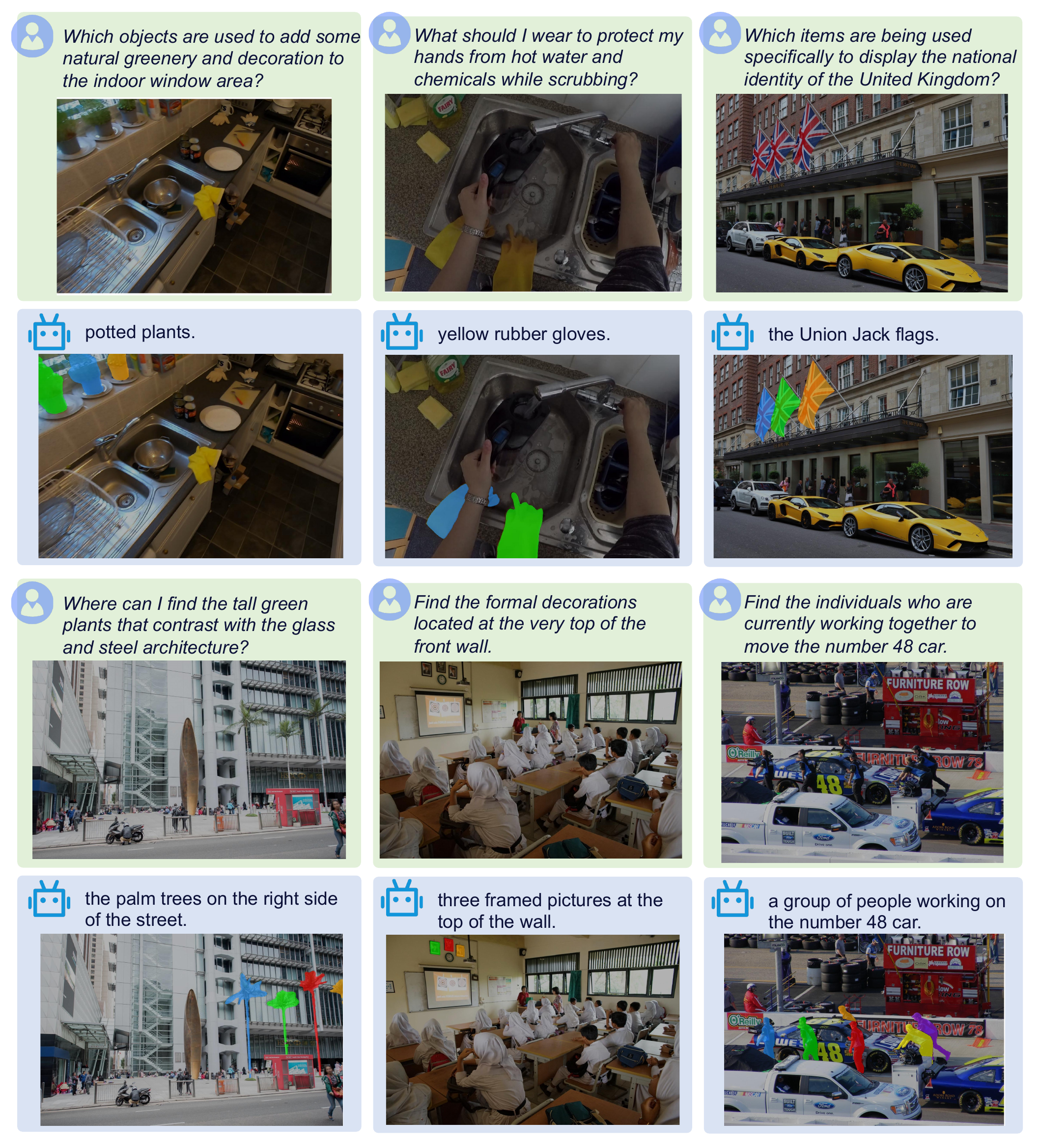}
\caption{Visualization of InstructSAM on instruction-based instance segmentation.}\label{fig: vis1}
\end{figure*}

\noindent\textbf{Instruction-based Instance Segmentation.}
Fig.~\ref{fig: vis1} presents qualitative results of InstructSAM on the Instruction-based Instance Segmentation task. Given a natural-language query that may include multiple attributes (e.g., object category, spatial relations, and contextual constraints), InstructSAM is able to accurately parse the instruction, localize the referred instance, and produce a precise instance-level segmentation mask. As illustrated, the model robustly handles complex and compositional descriptions, distinguishing the target object from visually similar distractors and cluttered backgrounds. These results demonstrate that InstructSAM effectively bridges language understanding and fine-grained mask generation, enabling reliable instance segmentation guided directly by user instructions.

\begin{figure*}[h!]
\centering
\includegraphics[width=\linewidth]{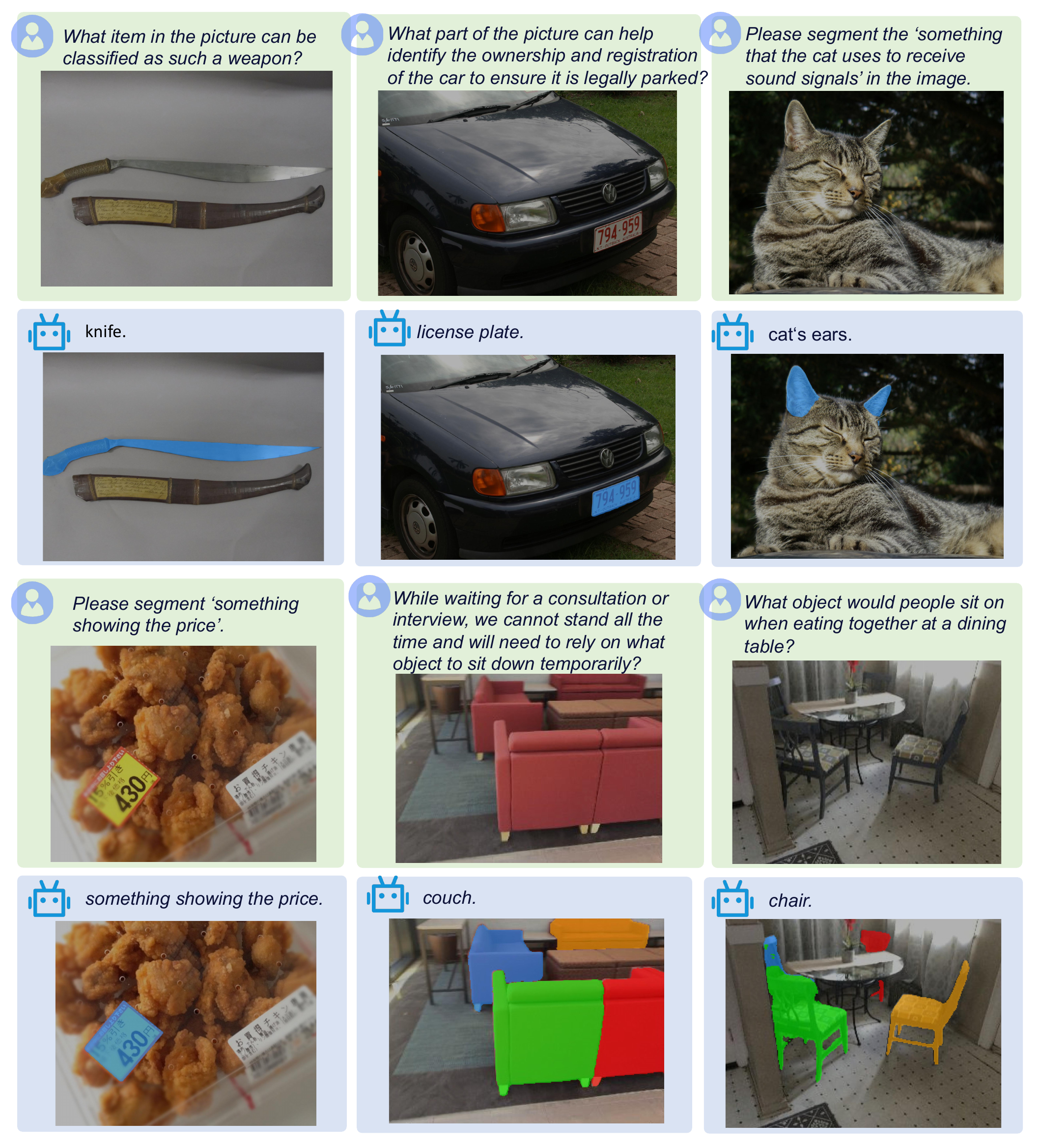}
\caption{Visualization of InstructSAM on reasoning segmentation.}\label{fig: vis3}
\end{figure*}

\noindent\textbf{Reasoning Segmentation.} 
Fig.~\ref{fig: vis3} visualizes InstructSAM's performance on the reasoning segmentation task. In this setting, the query cannot be resolved by category recognition alone; instead, the model must perform multi-step reasoning, such as identifying functional parts (e.g., ``the part used to receive sound signals''), inferring intent or affordances (e.g., ``something to sit on temporarily''). As shown, InstructSAM produces masks that align with the inferred evidence regions, demonstrating strong commonsense grounding and the ability to translate reasoning outcomes into precise, localized segmentations.

\begin{figure*}[h!]
\centering
\includegraphics[width=\linewidth]{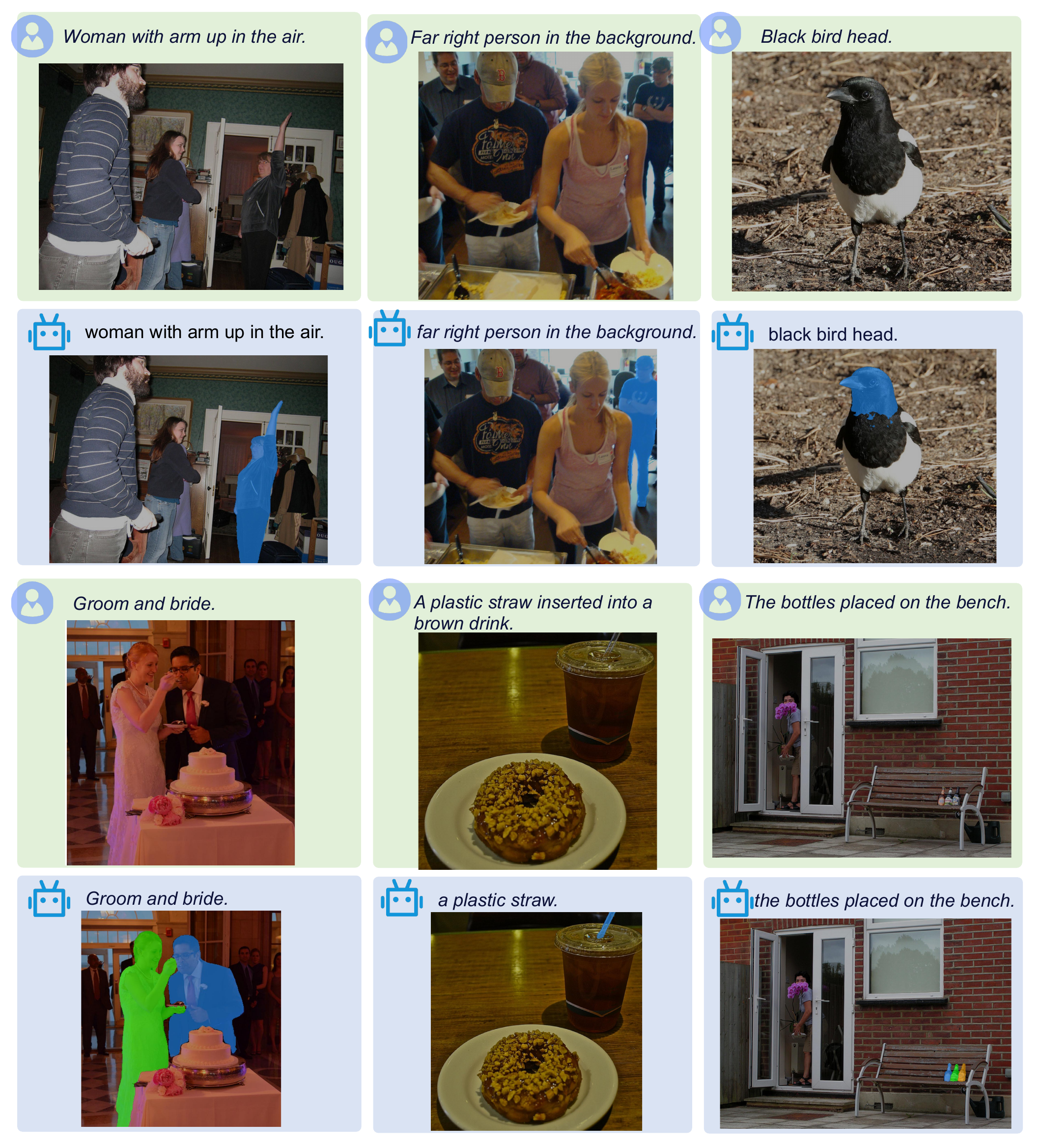}
\caption{Visualization of InstructSAM on referring segmentation.}\label{fig: vis2}
\end{figure*}

\noindent\textbf{Referring Segmentation.}
Fig.~\ref{fig: vis2} illustrates the visualization results of InstructSAM on the Referring Segmentation task. In this setting, the model receives a referring expression that may describe the target region through appearance cues, relative position, or contextual relations (e.g., ``far right person in the background'' or ``the bottles placed on the bench''). As shown in the examples, InstructSAM can accurately ground the expression in the scene and segment the corresponding region with clear boundaries, even under challenging conditions such as crowded scenes, small objects, and background clutter. These qualitative results further verify the strong language grounding ability of InstructSAM and its robustness in producing fine-grained masks aligned with diverse referring expressions.

\section{Further Discussions}\label{secA4}

Although InstructSAM demonstrates strong performance on instruction-based segmentation, several limitations remain. \textbf{First}, the current version focuses on image inputs, primarily because constructing large-scale, high-quality instruction-mask supervision for videos is substantially more challenging. In video scenarios, multi-instance interactions, temporal correspondence, and frame-level mask consistency make data annotation and automatic data generation more complex, while also increasing the risk of hallucinated or inconsistent instructions. Extending the training and evaluation to video data, for example through joint training with video segmentation datasets, could further enhance temporal reasoning and mask consistency, potentially improving segmentation robustness in dynamic scenes. \textbf{Second}, effectively integrating segmentation datasets with existing large-scale conversational instruction involving complex reasoning remains an open challenge. 
Naive joint training can degrade performance on certain segmentation benchmarks, indicating non-trivial interference between datasets and objectives. Designing principled co-training strategies, such as data balancing schemes that incorporate conversational supervision while preserving or improving segmentation fidelity, is therefore crucial for further advancing instruction-following segmentation models.

\bibliographystyle{unsrtnat}
\bibliography{main}

\end{document}